\documentclass[journal]{IEEEtran}
\ifCLASSINFOpdf
\else
\fi
\usepackage{amsmath}
\usepackage{amssymb}
\usepackage[pdftex]{graphicx}
\usepackage{algorithm}
\usepackage{algorithmicx}
\usepackage{algpseudocode}
\usepackage{color}
\usepackage{multirow}
\usepackage{makecell}
\usepackage{cite}

\usepackage{subfigure}
\usepackage{threeparttable}
\usepackage{stfloats}

\newtheorem{remark}{Remark}
\begin{document}
%
\title{Reducing Learning Difficulties: One-Step Two-Critic Deep Reinforcement Learning for Inverter-based Volt-Var Control in Active Distribution Networks}
%
%

\author{Qiong~Liu,
        Ye~Guo,
        Lirong~Deng,
        Haotian~Liu,
        Dongyu~Li,
        Hongbin~Sun,
        and Wenqi Huang
\thanks{This work was supported in part by the National Key R\&D Program of China (2020YFB0906000, 2020YFB0906005).}
\thanks{Qiong Liu, Ye Guo are with the Tsinghua-Berkeley Shenzhen Institute, Tsinghua University, Shenzhen, 518071, Guangdong, China, e-mail: guo-ye@sz.tsinghua.edu.cn.}
\thanks{Lirong Deng is with the Department of Electrical Engineering, Shanghai University of Electric Power, Shanghai, 200000, China}
\thanks{Haotian Liu, Hongbin Sun are with the State Key Laboratory of Power Systems, Department of Electrical Engineering, Tsinghua University, Beijing 100084, China}
\thanks{Donyu Li is with the School of Cyber Science and Technology, Beihang University, Beijing, 100191, China.}
\thanks{Wenqi Huang is with  the  Digital Grid Research Institute, China Southern Power Grid, Guangzhou, China.}
\thanks{Manuscript received XX, 2022; revised XX, 2022.}}

%
%

\markboth{IEEE TRANSACTIONS ON SMART GRID, ~VOL.~XX, NO.~X, April 2022}%
{Shell \MakeLowercase{\textit{et al.}}: Bare Demo of IEEEtran.cls for IEEE Journals}
%



\maketitle

\begin{abstract}
A one-step two-critic deep reinforcement learning (OSTC-DRL) approach for inverter-based volt-var control (IB-VVC) in active distribution networks is proposed in this paper.
First, the problem of IB-VVC is formulated as a one-step Markov decision process, which reduces the difficulties of the DRL learning task.
Correspondingly, a one-step actor-critic DRL scheme is designed, which has a simpler structure and avoids the problem of Q-value over-estimation.
Second, considering two objectives of VVC: minimizing power loss and eliminating voltage violations, we utilize two critics to approximate the rewards of two objectives separately, which reduces the difficulties of the approximation tasks of each critic.
OSTC-DRL under the simper structure improves the approximation accuracy of critics, accelerates the convergence process, and improves the control  performance.
The OSTC-DRL approach cooperates well with any  actor-critic DRL algorithms for the centralized IB-VVC problem, and two centralized DRL algorithms were taken as examples.
The multi-agent OSTC-DRL approach is also developed and applied to the decentralized IB-VVC problem.
Extensive simulation experiments show that the two proposed OSTC-DRL algorithms require fewer iteration times and return better results than the recent DRL algorithms, and the multi-agent OSTC-DRL algorithms work well for decentralized IB-VVC problems.

\end{abstract}



%
\IEEEpeerreviewmaketitle

\section{Introduction}
%
%
%
%
\IEEEPARstart {O}n our efforts toward a carbon-neutral society, more and more distributed energy resources will be integrated into active distribution networks (ADNs). The increasing penetration of distributed generations (DGs) poses new challenges to voltage regulation.
To tackle this problem, Volt-Var control (VVC) will be more and more important for active distribution networks. VVC optimizes the output of the reactive power resources to eliminate voltage violations and minimize power loss.
Most DGs are inverter-based energy resources (IB-ERs) that are possible to provide reactive power support rapidly. There is an increasing interest to utilize these resources to achieve the VVC \cite{haiyueRobust, Tang_realtime}.

Recent methods for inverter-based VVC (IB-VVC) problems can be divided into model-based and data-driven. Model-based methods solve the IB-VVC problems based on a reliable model of the ADN. However, such a reliable model may be difficult to acquire for distribution system operators \cite{PhysRevE69025103}.
As an alternative solution, data-driven methods learn optimal actions from measurements directly.  In data-driven methods, deep reinforcement learning (DRL) methods are intensively studied.
Generally, actor-critic DRL algorithms are usually implemented to deal with IB-VVC problems \cite{liu2021bi,DIcao_MADRL_PV, kou2020safe, Gao_Consensus, liu2021online}.
Actor-critic algorithms train both the state-action value network (also named as critic) and the policy network (also named as actor).
They reduce the variance, improve data efficiency, and accelerate the learning process \cite{sutton2018reinforcement}.
Meanwhile, DRL algorithms can achieve real-time decision-making by shifting the computational expense from online optimization to offline training.
However, they may still suffer optimality problems especially in minimizing power loss.
The learning process may be slow, and the convergence tends to be unstable. 

	

From the perspective of the general DRL field, the intrinsic reason for the above problems is the unavoidable estimation error of critic networks including overestimation or underestimation \cite{thrun1993issues}.
The overestimation error is accumulated when using the temporal difference learning method to train critic networks \cite{fujimoto2018addressing}. It may worsen the performance of the policy network or even lead to divergent behavior.
The underestimated bias will not be explicitly accumulated through the policy update but may still degrade performance \cite{pan2020softmax, ciosek2019better}.
Deep deterministic policy gradient (DDPG) uses a replay buffer and soft target updates to alleviate the overestimation of critics \cite{lillicrap2016continuous}.
Twin delayed deep deterministic policy gradient (TD3) \cite{fujimoto2018addressing} and soft actor-critic (SAC) \cite{haarnoja2018softsac} mitigates the overestimation bias and its accumulation by the technology of clipped double-Q learning.
Nevertheless, clipped double-Q learning leads to an underestimation bias, which also degrades performance \cite{pan2020softmax, ciosek2019better}.
Applying TD3 or SAC algorithms to VVC problems directly may encounter similar problems.

From the property of VVC problems, two objectives of VVC: minimizing power loss and eliminating voltage violations, brings additional difficulties to the learning process of DRL algorithms. Generally, the reward of DRL is designed as the combination of power loss and voltage violation rate \cite{haotian_Two_Stage,Gao_Consensus}.  The weight ratio between power loss and voltage violations needs to be tuned carefully, otherwise, the performance of DRL would degrade \cite{Wangwei-Safe-Off-Policy,zhang2020deep}.
To alleviate the problem, a Lagrangian relaxation method is introduced to tune the weight ratio between power loss and voltage violations in the reward design \cite{Wangwei-Safe-Off-Policy}.
To simply the learning task, the reward is the negative voltage violation rate when voltage violations appear, and the negative of the power loss when no voltage violation appears \cite{zhang2020deep}.
The critic network only needs to learn the penalty of voltage violations when voltage violations appear. 
The above reward design methods still mix the two objectives in one critic, and degrade the estimation accuracy.
To address the estimation error of the critic network, papers \cite{yan2020real, sun2021two} derive the critic based on the power flow model.
It avoids the inaccuracy estimation of critic value directly. However, the methods need a reliable model of ADNs, and the error of the model would degrade the VVC performance.

%

Given the literature review above, we observe that two main problems may degrade the performance of DRL for IB-VVC in ADNs: 1) The intrinsic problem is the estimation errors of critic networks including the overestimation and underestimation errors; 2) The two objectives of VVC bring additional difficulties to the learning process of critic networks in DRL algorithms.
To address the two problems, we propose a one-step two-critic DRL (OSTC-DRL) approach for IB-VVC in ADNs.
For the first problem, we formulate the IB-VVC as a one-step Markov decision process (MDP), 
and design a one-step actor-critic DRL scheme to solve the one-step MDP.
For the second problem, we use two critic networks to  approximate the rewards of power loss and voltage violation rate separately.
Based on the OSTC-DRL approach, we design two kinds of centralized DRL algorithms which are OSTC with a deterministic policy (OSTC-DP) derived from DDPG and OSTC soft actor-critic (OSTC-SAC) derived from SAC.
For decentralized VVC, we extend the OSTC-DRL approach to a multi-agent OSTC-DRL approach and design two multi-agent DRL algorithms which are multi-agent OSTC-DP and multi-agent OSTC-SAC.
Our proposed approach is simple, stable, and efficient. Compared with the existing DRL-based VVC algorithms, the main contributions of this paper and the technical advancements are summarized as follows:

\begin{itemize}
\item[1.] We propose an OSTC-DRL approach for IB-VVC in ADNs that accelerates the convergence rate and improves control performance by reducing the difficulty of DRL learning tasks and decreasing the approximation difficulty of each critic. It is a simple yet effective DRL approach for IB-VVC.

\item[2.]  We analyze that general DRL approaches like DDPG or SAC for IB-VVC problems complicate the problems and would lead to overestimation or underestimation issues due to the incompatible between the approaches and the problems, and  the proposed OSTC-DRL approach addresses those issues by learning the policy to maximize the recent reward rather than the infinity discounted accumulated reward.

\item[3.] We analyze that the two objectives of IB-VVC would increase the learning difficulties of one critic, and the proposed OSTC-DRL approach addresses the issue by using two critics to learn the two objectives separately.


\item[4.] We extend the OSTC-DRL approach to the multi-agent OSTC-DRL approach for decentralized IB-VVC. 
Multi-agent OSTC-DRL algorithms have a similar performance to the centralized OSTC algorithms even when each actor executes based on the local one bus information. 
The associated codes in this paper will be shared on Github.
 \end{itemize}

\section{Problem Formulation} \label{Problem_Formulation}

Generally, the problem of IB-VVC is formulated as an MDP. MDP is designed for formulating long horizon decisions or multi-period optimization problems. However, the IB-VVC problem can be formulated as a single-period optimization problem.
Using MDP to formulate the problem is feasible but it increases the complexity of optimization tasks.

For a single-period problem, it would be better to formulate the problem as a one-step MDP, also known as the contextual bandit \cite{majzoubi2020efficient}. 
The one-step MDP is defined by a tuple $(\mathcal{S},\mathcal{A},\mathcal{R})$.
At each time step, the agent observes a state $s \in \mathcal{S}$, and selects actions $a \in \mathcal{A}$ based on its policy $\pi: \mathcal{S} \rightarrow \mathcal{A}$, receiving a reward $r \in \mathcal{R}$. For the one-step MDP, the next state is not included in the process.

For the IB-VVC problems, state space, action space, and reward function are defined for one-step MDP as follows: 
\begin{itemize}
\item[1)] State: 
The state is  $s = (P, Q, V,Q_{G})$.
$Q_{G}$ is the vector of reactive power generations of controllable devices, such as IB-ERs and SVCs.
Since this paper focuses on the IB-VVC, for simplicity, we assume the topology of the ADN does not change in the control process, so the topology information is not added to the state \cite{haotian_Two_Stage}.
Adding $Q_{G}$ is to reflect the working condition of control variables in ADNs. Otherwise, we cannot obtain load reactive power $Q_D$ and inverter reactive power $Q_{G}$ with only the nodal reactive power $Q$.


\item[2)] Action: The action is implemented after the agent observe the state $s$. The action is $ a= Q_{G}$, where  $Q_{G}$ are the reactive power outputs of all IB-ERs and SVCs, and the range $\left|Q_{G}\right| \leq \sqrt{S_{G}^{2}-\overline{P_{G }}^{2}}$ for IB-ERs and $\underline{Q_{G}} \leq Q_{G} \leq \overline{Q_{G}}$ for SVC \cite{Two_Timescale_yang,Stochastic_Kekatos}. $\overline{P_{G}}$ is  the upper limit of active power generation of IB-ERs, and $\overline{Q_{G}}$ and $\underline{Q_{G}}$ the upper and bottom limit of reactive power generation of SVCs.


\item[3)] Reward:  We assume the new state $s^\prime$ can be obtained after executing action $a$ immediately. Reward is calculated based on $s^\prime$.
VVC problems have two objectives: minimizing active power loss, and eliminating voltage violations. Hence, the reward consists of two terms: the negative of active power loss $r_{p}$, and the negative of voltage violation rate $r_{v}$. The reward of power loss $r_{p}$ is defined as:
 \begin{equation}
 r_{p} =  - \sum P^\prime.
 \end{equation}
Similar to \cite{zhang2020deep, sun2021two}, the reward of voltage violation $r_{v}$ is defined as
\begin{equation}
r_v=-\sum \left[\max \left(V^\prime-\bar{V}, 0\right)+\max \left(\underline{V}-V^\prime, 0\right)\right].
\end{equation}

\end{itemize}

\section{One-Step Two-Critic DRL}
\begin{figure}[!t] 
\centering
\includegraphics[width=3in]{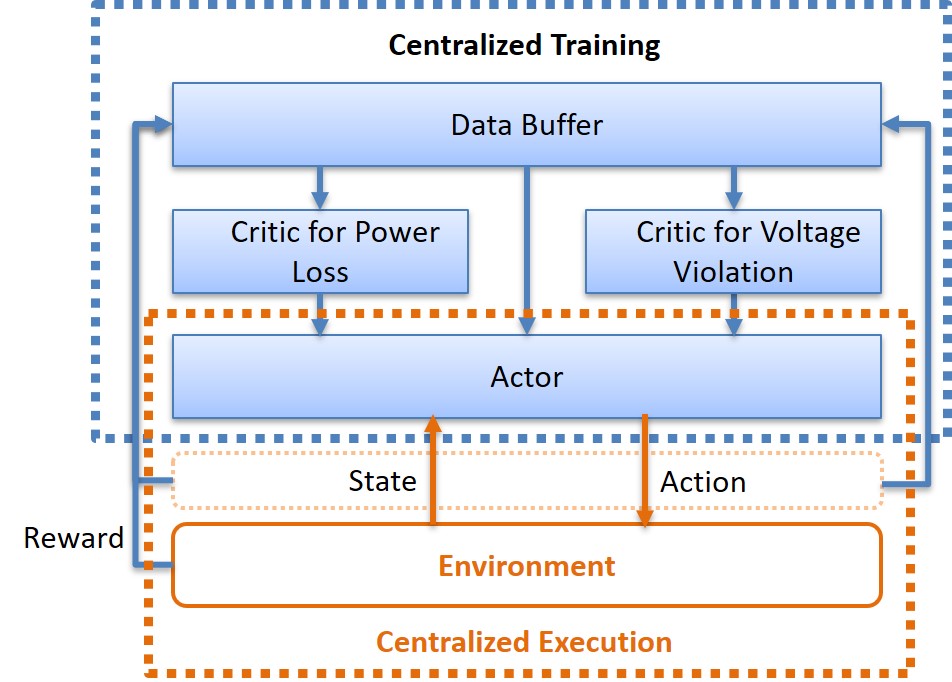}
\caption{The framework of one-step two-critic deep reinforcement learning.}
\label{CTCE}
\end{figure}

This section introduces details of the OSTC-DRL approach. We first propose a one-step DRL scheme to learn from the one-step MDP data, and then utilize the two-critic to learn the reward of power loss and voltage violations separately. Fig. \ref{CTCE} shows the framework of the one-step two-critic DRL approach.

\subsection{One-Step Actor-Critic DRL} \label{SS-AC}

Generally, actor-critic DRL algorithms, such as DDPG, TD3, and SAC are used to solve the IB-VVC problem. However, it complicates the problem and leads the overestimation or underestimation issues because of the following reasons:
\begin{itemize}
\item[1.] Actor-critic DRL algorithms usually solve long horizontal decision problems. However, IB-VVC is a single-period optimization problem.
Directly applying those DRL algorithms in IB-VVC problems complicates the problem. It may need more time and more data in the learning process.
\item[2.] The overestimation issue occurs because actor-critic DRL algorithms like DDPG use temporal difference learning to update the critic network. The temporal difference learning is  
\begin{equation}
    y_t = r_t + \gamma Q_\phi(s_{t+1},a_{t+1}), \   a_{t+1} \sim  \pi_{\theta}(s_{t+1}).
\end{equation}
The overestimation error occur in the term $Q_\phi(s_{t+1},a_{t+1})$ and the overestimation error would accumulate in the temporal difference learning \cite{fujimoto2018addressing}. 
\item[3.] To address the overestimation issue,  clipped double-Q learning is proposed in TD3 or SAC. Clipped double-Q learning is 
\begin{equation}
    y_t = r_t + \gamma  \min_{i = 1,2} Q_{\phi_i}(s_{t+1},a_{t+1}), \   a_{t+1} \sim  \pi_{\theta}(s_{t+1}).
\end{equation}
The term $\min_{i = 1,2} Q_{\phi_i}(s_{t+1},a_{t+1})$ address the overestimation issue but leads to the underestimation issue. The underestimation issue would not accumulate in the learning process, but still degrade the learning performance \cite{pan2020softmax, ciosek2019better}.
\end{itemize}
To address the problems above, we design the one-step DRL approach where the objective of one-step DRL is to learn a policy $\pi$ that maximizes the expected reward,
\begin{equation}
	\pi^* = \arg \max_\pi E_{a \sim \pi} ( r(s,a) ),
\end{equation}
where $r = r_p +c_v r_v$, and $c_v$ is a constant. 
The learning task is simpler compared with the ordinary DRL algorithm because the ordinary DRL algorithms maximizes the expected infinite-horizon discounted accumulated reward $E_{a \sim \pi} (\sum_{t=0}^\infty \gamma r_t(s,a) )$, where $\gamma<1$ is a discount factor.  Meanwhile, the term $Q_{\phi_i}(s_{t+1},a_{t+1})$ is not in the process of learning critic networks, thus avoiding the overestimation or underestimation issue successfully.  


Actor-critic DRL algorithms use the state-action value function $Q(s, a)$  (known as the critic) to guide the update of the policy $\pi$ (known as the actor).
For one-step actor-critic DRL, the critic $Q(s,a)$ is the expected reward for state $s$ and action $a$,	
	\begin{equation}\label{policy_gradient}
Q^{\pi}( s, a)=\underset{a \sim \pi}{E}\left[  r \mid s,a \right].
\end{equation}

The aim of the actor is selecting an action to maximize its critic value $Q(s,a)$: 
\begin{equation}
\pi(a|s) = \arg \max_{a} Q\left(s, \pi(a|s)\right).
\end{equation}

In practical implementation, both the critic $Q\left(s, a \right)$ and the actor $\pi(a|s)$ are approximated by neural networks $Q_\phi$ and $\pi_\theta$ with parameters of $\phi$ and $\theta$.  $Q_\phi(s,a)$ is learned by minimizing the MSE loss: 
\begin{equation}\label{MSE}
 L_Q(\phi)  = \frac{1}{|B|} \sum_{\left(s, a, r \right) \in B}\left(Q_{\phi}(s, a)-r\right)^{2},
\end{equation}
where  $B$ is a mini-batches data $\left(s, a, r \right)$ sampling from  the replay buffer, and $|B|$ is the number of samples in the batch.

The actor $\pi_\theta$ is learned by maximizing the loss function:
\begin{equation}\label{pg}
L_\pi({\theta}) = \frac{1}{|B|} \sum_{s \in B} Q_{\phi}\left(s, \mu_{\theta}(s)\right).
\end{equation}

%
%
%
%


\begin{remark}
The one-step DRL scheme is the simplest version of DDPG or TD3.  Setting $\gamma = 0$ for DDPG or TD3 would derive the scheme directly.  After setting $\gamma = 0$, the target network, clipped double-Q learning, delayed policy updates, target policy smoothing, and the soft updating of the target network is not necessary for DDPG or TD3.
Compared with DDPG with 4 neural networks, and TD3 with 6 neural networks, the scheme only needs 2 neural networks. 
However, the scheme would have better performance because it is designed for the IB-VVC problem specifically.
\end{remark}

\subsection{Two-Critic Scheme for One-Step Actor-Critic DRL}



\begin{figure}[!t]
\centering 
\subfigure[Two-bus system.]{
\includegraphics[width=2in]{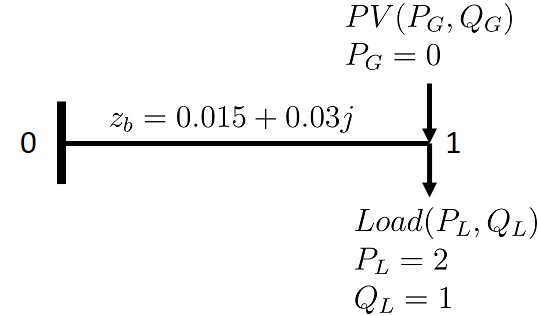}}
\subfigure[The functions $V(Q_G)$, $P_{loss}(Q_G)$.]{
\includegraphics[width=2in]{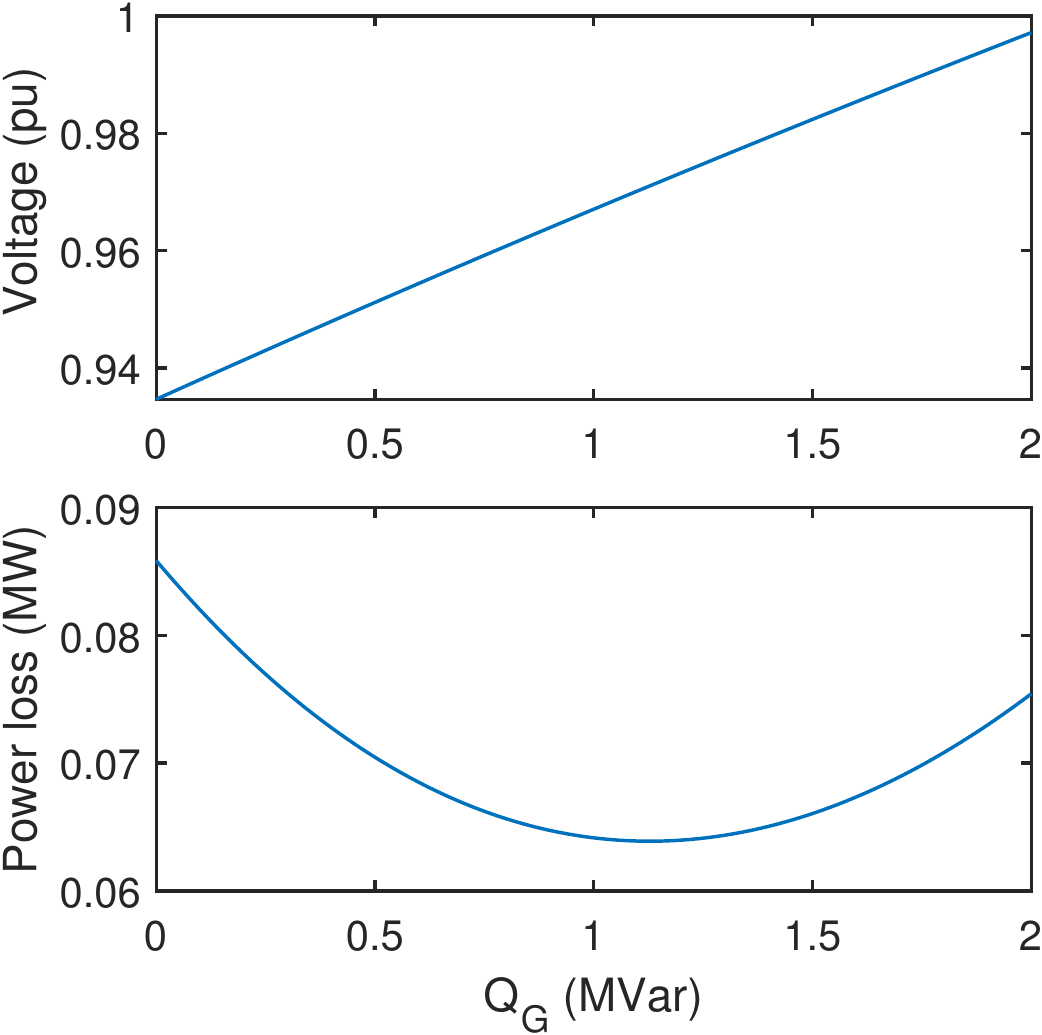}}
\caption{The relationship between voltage $V$, power loss $P_{loss}$ and reactive  power generation $Q_G$ for a two-bus system.}
\label{two-bus}
\end{figure}

VVC has two objectives: minimizing power loss, and eliminating voltage violations. For such a problem with multiple objectives, existing DRL methods integrate them into one objective and use one critic network to learn it.
Neural networks indeed have the universe approximation capability to approximate any continuous function with arbitrarily small error with enough number of hidden nodes theoretically \cite{chen1995universal}.
However, using one critic network to learn the integration of two objectives may degrade its approximation accuracy and convergence speed in the learning process because of the following properties of VVC problems:
	\begin{itemize}
	\item[1)] The reward for active power loss $r_p$ and the reward for voltage violation rate $r_v$ have different properties. As shown in Fig. \ref{two-bus},  the relationship between reactive power injection $Q_G$ and voltage $V$  at ADNs is close to linear, whereas the relationship between reactive power injection $Q_G$ and power loss $P_{loss}$ is strongly nonlinear. If we use one critic network to approximate the reward containing two objectives, the critic network may have to mix the two functions.
	In addition, the data in the reply buffer is dynamic and does not satisfy the independent-identical-distribution condition, which increases the difficulty further. 
	\item[2)] Eliminating voltage violations is more important than minimizing power loss, so $c_v$ should be set large enough to penalize voltage violations. However, the performance of the critic may suffer from the numerical stability problem for a large $c_v$ \cite{zhang2020deep}.
	\item[3)] Difficulties in learning the function of power loss and voltage violation may be different. The difficult task costs more time or more data, and the approximation accuracy may be lower compared with the easier task. One critic may mix the two tasks, and the worse performance of the difficult task maybe dominate in the learning process.
	
	\end{itemize}


To address the problems above, we use two critic networks to approximate the two objectives separately. 
The two-critic scheme decreases the approximation difficulties of each critic, thus having a faster convergence rate
and a small approximation error. Correspondingly, the reward stored in MDP is designed as $r=(r_p,r_v)$.
	

The critic  for power loss $Q_p(s,a)$ and voltage violations $Q_v(s,a)$ are	
\begin{equation}\label{two_Q}
\begin{split}
Q_p^{\pi}( s, a)&=\underset{a \sim \pi}{E}\left[  r_p \mid s,a \right] \\
Q_v^{\pi}( s, a)&=\underset{a \sim \pi}{E}\left[  r_v \mid s,a \right].
\end{split}
\end{equation}
	 
In real application, the two critics $Q_p\left(s, a \right)$ and $Q_p\left(s, a \right)$ are approximated by two neural networks $Q_{\phi_p}$ and $Q_{\phi_v}$ with parameters of $\phi_v$ and $\phi_v$.  $Q_{\phi_p}(s,a)$ and $Q_{\phi_v}(s,a)$ are learned by minimizing the MSE losses,	
\begin{equation}\label{two_QMSE}
\begin{split}
L_{Q_p}(\phi_p) &= \frac{1}{|B|} \sum_{\left(s, a, r_p \right) \in B}\left(Q_{\phi_p}(s, a)-r_p\right)^{2}\\
L_{Q_v}(\phi_v) & = \frac{1}{|B|} \sum_{\left(s, a, r_v \right) \in B}\left(Q_{\phi_v}(s, a)- c_v r_v\right)^{2}.
\end{split}
\end{equation}

The network actor is updated by maximizing the loss function,
\begin{equation}\label{twopg}
L_{\pi}({\theta}) =   \frac{1}{|B|} \sum_{s \in B}  \left( Q_{p_{\phi}}\left(s, \pi_{\theta}(s)\right)  + Q_{v_{\phi}}\left(s, \pi_{\theta}(s)\right)   \right).
\end{equation}
%

\begin{algorithm}[!t]
  \caption{One-Step Two-Critic with Deterministic Policy (OSTC-DP) algorithm} \label{OSTC-DP-algorihtm}
  \begin{algorithmic}[1]
  \Require
  Initial policy parameters $\theta$, Q-function parameters $\phi_p$, $\phi_v$,  empty replay buffer $\mathcal{D}$;
\For {each environment step}
\State  Observe  state $s$, and select action  $a=$ 
\Statex \quad \ \   $clip (\pi_{\theta}(s) + \epsilon, a_{Low}, a_{High})$, where $\epsilon \sim \mathcal{N}$;
\State Execute $a$ in the environment, and observe reward
\Statex \quad \ \  $r_p, r_v$;
\State Store $(s,a,r_p, r_v)$ in replay buffer;
\If {it's time to update} 

\For {$j$ in range (how many updates)} 

 \State Randomly sample a batch of transitions, 
\Statex \qquad \qquad \  $B={(s,a,r_p,r_v)}$ from $\mathcal{D}$;
 
\State Update  $Q_p$ and $Q_v$ by one step of
\Statex \qquad \qquad \ \   gradient descent using
$$
\nabla_{\phi_p} \frac{1}{|B|} \sum_{\left(s, a, r_p \right) \in B} \left(Q_{\phi_p}(s, a)- r_p\right)^{2}
$$
$$
\nabla_{\phi_v} \frac{1}{|B|} \sum_{\left(s, a, r_v \right) \in B} \left(Q_{\phi_v}(s, a)- r_v\right)^{2}
$$

\State Update policy by one step of gradient ascent  \Statex \qquad \qquad \ \  using 
\begin{equation}\nonumber
\begin{split}
\nabla_{\theta} \frac{1}{|B|} \sum_{s \in B}  \big( Q_{p_{\phi}}\left(s, \pi_{\theta}(s)\right) \\ + c_v Q_{v_{\phi}}\left(s, \pi_{\theta}(s)\right)   \big)
\end{split}
\end{equation}

\EndFor
\EndIf
\EndFor
\end{algorithmic}
\end{algorithm}

OSTC-DRL is compatible well with any off-policy actor-critic algorithms. For example, we design an OSTC-DP derived from DDPG to show the approach is compatible well with deterministic policies. Algorithm \ref{OSTC-DP-algorihtm} provides the detail of the OSTC-DP.
We also design OSTC-SAC derived from SAC to show the approach is compatible well with stochastic policies.
To obtain OSTC-SAC, we need to make little modifications in steps 2 and 9 in Algorithm \ref{OSTC-DP-algorihtm}. 
In step 2, we need to replace the deterministic policy $\pi_{\theta}(s)=\tanh \left(\mu_{\theta}(s)\right)$ as the stochastic policy $\pi_{\theta}(s, \xi)=\tanh \left(\mu_{\theta}(s)+\sigma_{\theta}(s) \odot \xi\right), \quad \xi \sim \mathcal{N}(0, I)$, where $\mu_{\theta}$ , $\sigma_{\theta}$ are neural networks.
In step 9, we need to add the  entropy regularization term, $\nabla_{\theta} \frac{1}{|B|} \sum_{s \in B}  \big( Q_{p_{\phi}}\left(s, \pi_{\theta}(s)\right)  + c_v Q_{v_{\phi}}\left(s, \pi_{\theta}(s)\right) -  \alpha \log \pi_{\theta}\left(\pi_{\theta}(s) \mid s \right)$. The temperature $\alpha$ can be adjusted by minimizing the loss 
$L(\alpha) =  \frac{1}{|B|} \sum_{s \in B , a \sim \pi_\theta} [-\alpha \log \pi_\theta (a|s) - \alpha \mathcal{H}] $, where $\mathcal{H}$ is the entropy target. 

\section{Extending OSTC-DRL to Multi-agent OSTC-DRL for Decentralized IB-VVC}	

\begin{figure}[!t] 
\centering
\includegraphics[width=3.5in]{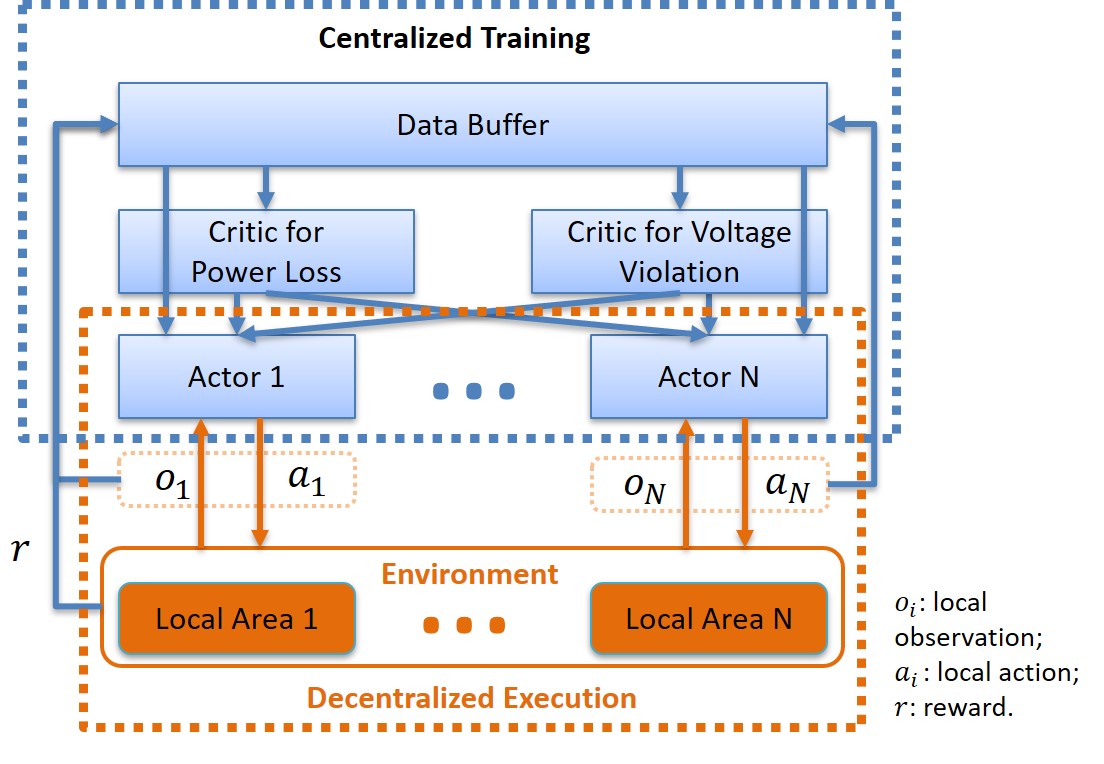}
\caption{The framework of multi-agent one-step two-critic deep reinforcement learning.}
\label{CTDE}
\end{figure}

The centralized OSTC-DRL approach requires massive real-time  communication and is fragile for single-point  failure.
For ADNs that each sub-area only can acquire local information in real-time, and to enhance the robustness against communication failures, we extend the OSTC-DRL to the multi-agent form for decentralized IB-VVC. It is based on a centralized training decentralized execution approach.
In off-policy DRL algorithms, both actor networks and critic networks are trained by the data sampling from the data buffer, so the interaction data stored in the data buffer with a delayed time has little influence in the training stage. In executing stage, each actor-network of the sub-area just requires real-time local observation to make the decision.  The framework of multi-agent OSTC-DRL is shown in Fig. \ref{CTDE}.

\subsection{Formulating VVC as One-step Markov Game}
Markov game is an extension of MDP for a multi-agent system. Similar to the one-step MDP, we formulate the IB-VVC as a one-step Markov game where multiple agents interact with the same environment. 
We use a tuple $\left(\mathcal{S},\left[\mathcal{O}_{i}\right]_{n},\left[\mathcal{A}_{i}\right]_{n}, \mathcal{R}\right)$ to describe a Markov game with $n$ agent, where
$s \in \mathcal{S}$ is the full state of environment, $o_i \in \mathcal{O}_{i}$ and  $a_i \in  \mathcal{A}_{i}$ are the local observations and actions for each agent $i$, and
$[r_p,r_v]^T \in \mathcal{R}$ are the reward functions relating power loss and voltage violation rate that are defined as $\mathcal{S} \times \mathcal{A}_{1} \times \dots \mathcal{A}_{N}  \times \mapsto \mathcal{R}_p, \mathcal{R}_v $. The goal of each agent is to maximize the expected reward $J = E_{a_i \sim \pi_i(o_i)}(r_p(s,a_1,a_2,\dots,a_n)+ c_v r_v(s,a_1,a_2,\dots,a_n) )$.
	
The definition of state space and the reward is the same as the centralized version in section \ref{Problem_Formulation}.
The local observations of each agent is $o_{i} =\left(P_{i}, Q_{i},  Q_{G,i},Q_{C,i} \right)$, where $o_{i}$ is the local information of of $i^{th}$ subarea. 
The selection of local observation space depends on the measurement conditions of each sub-area.  It  can be the measurement of one sub-area and its neighbor areas, or just one bus of the IB-ER or SVC installed. 
The local action of each agent is  ${a}_{i,t}=\left(Q_{G,i}, Q_{C,i}\right)$, where $Q_{G,i}, Q_{C,i}$ are the controllable reactive power resources in $i^{th}$ area, including IB-ERs and SVCs.

\subsection{Multi-Agent OSTC}

As shown in Fig. \ref{CTDE}, multi-agent OSTC-DRL contains two stages: centralized training and decentralized execution.
In the centralized training stage, the off-policy multi-agent OSTC-DRL algorithm samples data from the data buffer to train actors and critics. 
In the decentralized execution stage, the actor generates actions and then applies the actions to the environment. The actions generated by actors are based on real-time local measurements. The interaction data will be stored in the data buffer, and the data storage can be at a slow rate.

Since the data buffer can obtain the full state of ADNs and actions of each actor at a slow rate, one critic $Q(s,a_1,\dots, a_n)$ is enough to represent the state-action value of all actors.
Similar to the OSTC-DRL, we design two critics that are the critic of  power loss $Q_p(s,a_1,\dots, a_n)$ and the critic of voltage violation $Q_v(s,a_1,\dots, a_n)$,
\begin{equation}\label{two_Q_MA}
\begin{split}
Q_p^{\pi}( s, a_1,\dots, a_n)&=\underset{a_i \sim \pi_i}{E}\left[  r_p \mid s,a_1,\dots, a_n \right] \\
Q_v^{\pi}( s, a_1,\dots, a_n)&=\underset{a_i \sim \pi_i}{E}\left[  r_v \mid s, a_1,\dots, a_n  \right].
\end{split}
\end{equation}

The critic networks $Q_{\phi_p}(s,a_1,\dots, a_n)$ and $Q_{\phi_v}(s,a_1,\dots, a_n)$ are the same as the centralized OSTC-DRL, which are learned by minimizing the MSE losses,
\begin{equation}\label{two_QMSE_MA}
\begin{split}
L_{Q_p}(\phi_p) &= \frac{1}{|B|} \sum_{\left(s, a_1,\dots, a_n, r_p \right) \in B}\left(Q_{\phi_p}(s, a_1,\dots, a_n)-r_p\right)^{2}\\
L_{Q_v}(\phi_v) &= \frac{1}{|B|} \sum_{\left(s, a_1,\dots, a_n, r_v \right) \in B}\left(Q_{\phi_v}(s, a_1,\dots, a_n)- r_v\right)^{2}.
\end{split}
\end{equation}

The network actors are updated by the loss function,
\begin{equation}\label{twopg_MA}
\begin{split}
L_\pi(\theta_1, \dots \theta_n) = \frac{1}{|B|} \sum_{s \in B}  \big( Q_{p_{\phi}}\left(s,  \pi_{\theta_1}(o_1), \dots, \pi_{\theta_n}(o_n)\right) \\
 + c_v Q_{v_{\phi}}\left(s, \pi_{\theta_1}(o_1), \dots, \pi_{\theta_n}(o_n) \right)   \big).
\end{split}
\end{equation}

Algorithm \ref{MAOSTC_al} shows the details of  multi-agent OSTC-DP.
In step 6, we need to collect the $(s,a_1, \dots, a_n,r_p,r_v)$ of the ADNs at the same time. It is accessible because recently measurement devices can add a timestamp to the measured data, and then the data can be updated to the center with a slow time rate or a constant time interval decay. Of course, those data may be non-synchronized, additional methods \cite{alimardani2015distribution} can be used to preprocess the non-synchronized data. 

To obtain multi-agent OSTC-SAC, we need to make little modifications in steps 3 and 10 in Algorithm \ref{MAOSTC_al}. 
In step 3, we need replace the policy as the stochastic policy
$\pi_{\theta_i}( o_i, \xi)=\tanh \left(\mu_{\theta_i}(o_i)+\sigma_{\theta_i}(o_i) \odot \xi_i \right), \quad \xi_i \sim \mathcal{N}(0, I)$, where $\mu_{\theta_i}$ , $\sigma_{\theta_i}$ are neural networks.
In step 10, we need add the entropy regularization term to the critic networks, $  L_\pi(\theta_1, \dots \theta_n) =  \frac{1}{|B|} \sum_{s \in B}  \big( Q_{p_{\phi}}\left(s,  \pi_{\theta_1}(o_1), \dots, \pi_{\theta_n}(o_n)\right) 
 + c_v Q_{v_{\phi}}\left(s, \pi_{\theta_1}(o_1), \dots, \pi_{\theta_n}(o_n) \right)    
 -\sum_i^n  \alpha \log \pi_{\theta_i}\left(\pi_{\theta}(s) \mid s \right) \big)$. The temperature $\alpha$ can be adjusted by minimizing the loss 
$L(\alpha) = \frac{1}{|B|} \sum_{s \in B , a \sim \pi_\theta} [-\alpha \log \pi_\theta (a|s) - \alpha \mathcal{H}] $, where $\mathcal{H}$ is the entropy target. 
  We name the algorithm derived from OSTC-SAC as multi-agent OSTC-SAC. 
  
 Multi-agent OSTC-DP and multi-agent OSTC-SAC are flexible for the measurement conditions of ADNs. The algorithms are designed for the measurement conditions that each actor executes based on the local single bus information or sub-area information.

\begin{algorithm}[!t]
  \caption{Multi-agent One-Step Two-Critic with Deterministic Policies (Multi-agent OSTC-DP) algorithm}\label{MAOSTC_al}
  \begin{algorithmic}[1]
  \Require
  Initial policy parameters $\theta_1, \dots \theta_n$, Q-function parameters $\phi_1$, $\phi_2$,  empty replay buffer $\mathcal{D}$;
\For {each environment step}
\For {each agent  $i$ }
\State  Observe $o_i$, and select action  $a_i=$
\Statex  \qquad \quad  $clip (\pi_{\theta_i}(o_i) + \epsilon_i, a_{Low}, a_{High})$, where $\epsilon_i \sim \mathcal{N}$;
\State Execute $a_i$ in the environment;
\EndFor

\State  Collect reward $r_p$, $r_v$ of the ADN;
\Statex \quad \   Store  $(s,a_1,\dots,a_n,r_p, r_v)$ in replay buffer;

\If {it's time to update} 

\For {$j$ in range (how many updates)} 

 \State Randomly sample a batch of transitions, 
$B=$
\Statex \qquad \qquad \quad $(s,a_1,\dots,a_n,r_p, r_v)$ from $\mathcal{D}$;
 
\State Update  $Q_p$ and $Q_v$ by one step of gradient
\Statex \qquad \qquad \ \  descent using Equ. \eqref{two_QMSE_MA};

\State Update policies by one step of gradient ascent \Statex \qquad \qquad \ \   using Equ. \eqref{twopg_MA};
\EndFor
\EndIf
\EndFor
\end{algorithmic}
\end{algorithm}

\section{Simulation}
	Numerical simulation was conducted on 33-bus \cite{baran1989network} and 69-bus \cite{das2008optimal} test distribution networks to demonstrate the advantages of the proposed OSTC-DRL approach. In the 33-bus test distribution network, 3 IB-ERs of 2 MVar reactive power capacity and 1.5 MW active power were connected to buses 18, 22, and 25, respectively, and 1 SVC of 2 MVar reactive power capacity was connected to bus 33.   
	In the 69-bus test distribution network, 4 IB-ERs of 2 MVar reactive power capacity and 1.5 MW active power were connected to buses 6, 24, 45, and 58 respectively, and 1 SVC of 2 MVar reactive power capacity was connected to bus 14. 
	All load and generation levels were multiplied with the fluctuation ratio of one day with 96 points extracted from paper \cite{haotian_Two_Stage} and a $20\%$ uniform distribution noise to reflect the variance.  
	The voltage limits for all buses were set to be [0.95, 1.05].
	The algorithms were implemented in Python. The balanced power flow was solved by Pandapower \cite{pandapower2018} to simulate ADNs, and the implementation of the DRL algorithms used PyTorch. 

\subsection{Simulations for the Centralized OSTC-DRL Approach}\label{simulation_centralized}
	
We performed simulations to understand the contribution of each component: one-step and two-critic, and show the superiority of the proposed centralized OSTC-DRL approach.
We designed 3  classes of simulation experiments. 
\begin{itemize}
    \item[] \textbf{Deterministic policy:} 
    1) DDPG \cite{lillicrap2016continuous};
    2) One-step with deterministic policy (OS-DP) derived from the DDPG by applying the one-step DRL scheme;
    3) Two-critic with deterministic policy (TC-DP) derived from the DDPG by applying the two-critic technology;
    4) One-step two-critic DRL with deterministic policy (OSTC-DP) derived from the DDPG by applying the one-step DRL scheme and two-critic technology.
    \item[] \textbf{Stochastic policy:} 
    5) SAC \cite{haarnoja2018soft};
    6) One-step with stochastic policy (OS-SAC) derived from the SAC by applying the one-step DRL scheme;
    7) Two-critic with stochastic policy (TC-SAC) derived from the SAC by applying the two-critic technology;
    8) One-step two-critic DRL with stochastic policy (OSTC-SAC) derived from the SAC by applying the one-step DRL scheme and two-critic technology.
    \item[] \textbf{Model-based:}  9) Model-based optimization method with accurate power flow model. Model-based optimization was solved by recalling PandaPower with the interior point solver. 
\end{itemize}

\begin{table}[!t]
\renewcommand{\arraystretch}{1.3}
\caption{Parameter setting for the reinforcement learning algorithm}
\label{DRL_paramter}
\centering
\begin{tabular}{ c c c}
\hline
 \text { Algo. } & \text { Parameter } & \text { Value } \\
\hline 
\multirow{15}{*}{\text{Shared}} & \text { Optimizer } & \text { Adam } \\
& \text { Activation function } & \text { ReLU } \\
& \text { Number of hidden layers } & 2  \\
& Actor hidden layer neurons & \{512, 512 \} \\
& Critic hidden layer neurons & \{512, 512 \} \\
& \text { Batch size } &  128 \\
& \text { Replay buffer size } & $ 3 \times 10^{4}$ \\
& \text {Critic learning rate} &  $ 3 \times 10^{-4}$\\
& \text {Actor learning rate   } &  $ 1 \times 10^{-4}$\\
& Voltage violation penalty $c_v$ & 50\\
& Initial random step & 960\\
& Iterations per time step & 4\\

\hline 
 The deterministic policy
& Exploration Policy & $\mathcal{N}(0,0.1)$ \\
\hline
\multirow{3}{*}{\text { The stochastic policy }}
&Entropy target & $-\dim(\mathcal{A})$\\
& Temperature  learning rate & $ 3 \times 10^{-4}$ \\
\hline 
\end{tabular}
\end{table}

\begin{table*}[!ht]
\centering
\begin{threeparttable}
\renewcommand{\arraystretch}{1.3}
\caption{Quantified indices of the benchmarks in the final 50 episodes for centralized DRL algorithms}
\label{DRL_performance}
\centering
\begin{tabular}{cccccccc}
\hline 
 \multicolumn{2}{c}{\multirow{2}{*}{ Algorithm }} &  \multicolumn{2}{c}{Reward}  &  \multicolumn{2}{c}{$P_{\text {loss }} / \mathrm{MW}$}   & \multicolumn{2}{c}{VVR/p.u.$^{2}$}  \\
\cline { 3 - 8 } 
 & & 33-bus & 69-bus &  33-bus & 69-bus &  33 -bus & 69-bus \\
\hline
Model-based & MBO$^{1}$ & $-4.199$ & $-3.957$ & $4.199$ & $3.957$  & 0  & 0\\
\hline
\multirow{2}{*}{\makecell[c]{Deterministic \\ policy}} 
& DDPG & $-4.898$ & $-4.701$ & $4.898$ & $4.622$  & $4.882\text{e-}6$ & $1.580\text{e-}3$ \\
& OS-DP & $-4.867$ & $-4.867$ & $4.867$ & $4.778$  & $1.129\text{e-}5$ & $1.782\text{e-}3$ \\
& TC-DP & $-4.289$ & $-4.217$ & $4.288$ & $4.140$  & $2.642\text{e-}5$ & $1.533\text{e-}3$ \\
&OSTC-DP & $\textbf{-4.270}$ &  $\textbf{-4.161}$ & $4.268$ & $4.104$  & $3.988\text{e-}5$ & $1.137\text{e-}3$\\
\hline
\multirow{2}{*}{\makecell[c]{Stochastic\\ policy}} 
& SAC & $-4.548$ & $-4.491$& $4.548$ & $4.471$ &         $0$  & $3.859\text{e-}4$ \\
&OS-SAC & $-4.368$ &$-4.383$ & $4.357$ & $4.339$ & $2.00\text{e-}4$  & $8.800\text{e-}4$ \\
&TC-SAC & $-4.384$ &$-4.275$ & $4.373$ & $4.205$ & $2.184\text{e-}4$  & $1.396\text{e-}3$ \\
&OSTC-SAC & $\textbf{-4.272}$ &$\textbf{-4.191}$ & $4.268$ & $4.121$ & $9.205\text{e-}5$  & $1.402\text{e-}3$ \\

\hline
\end{tabular}
\begin{tablenotes}
\item[1] MBO is the model-based optimization method using an accurate ADN model, which can be seen as the optimal result. However, the accurate ADN model is not available in real applications.
\item[2] VVR is the daily accumulation voltage violation rate.
\end{tablenotes}
\end{threeparttable}
\end{table*}

\begin{figure}[!t] 
\centering
\includegraphics[width=3.5in]{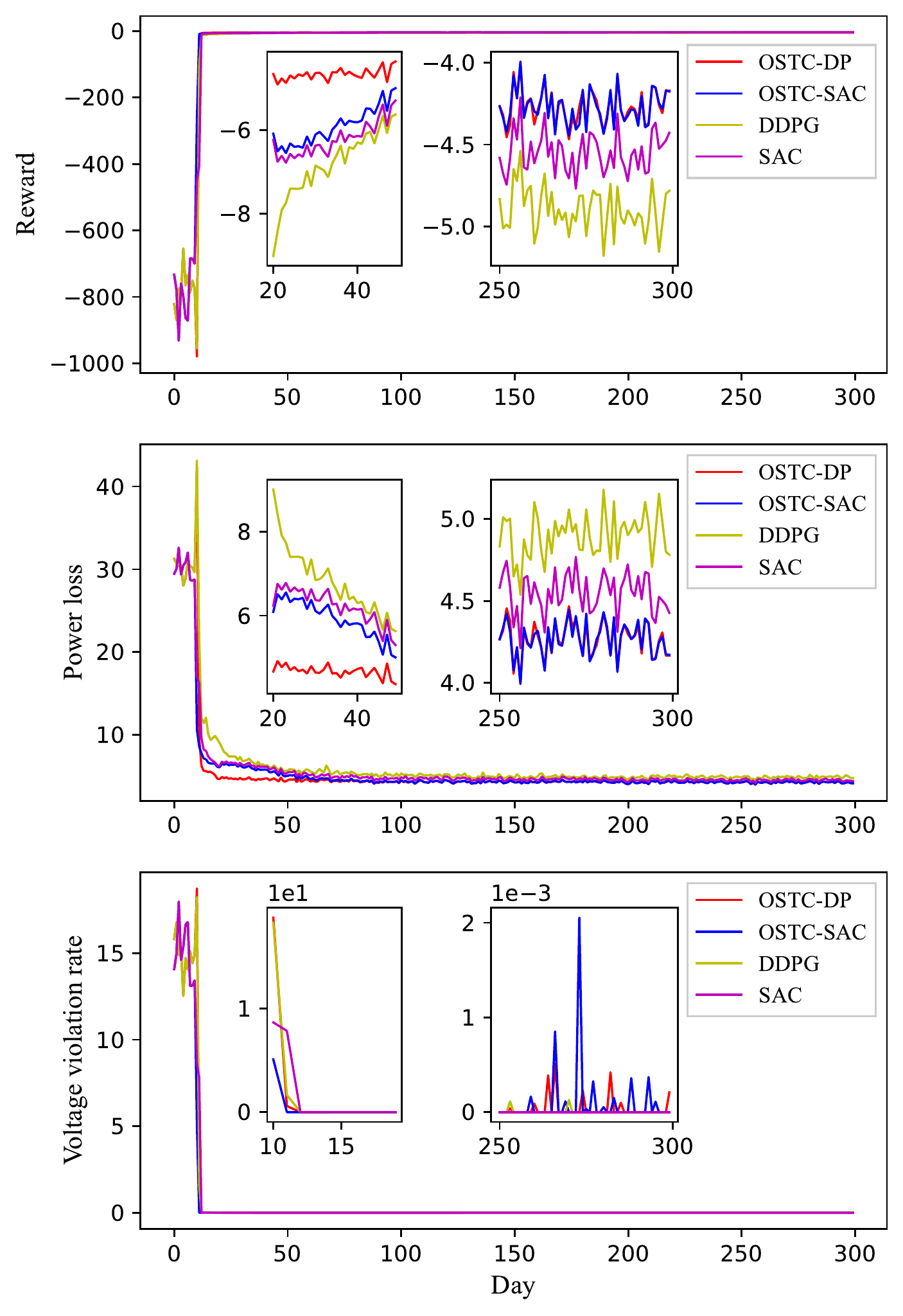}
\caption{Testing results of the training stage for the 33-bus test distribution network.}
\label{DRL_33}
\end{figure}

\begin{figure}[!t] 
\centering
\includegraphics[width=3.5in]{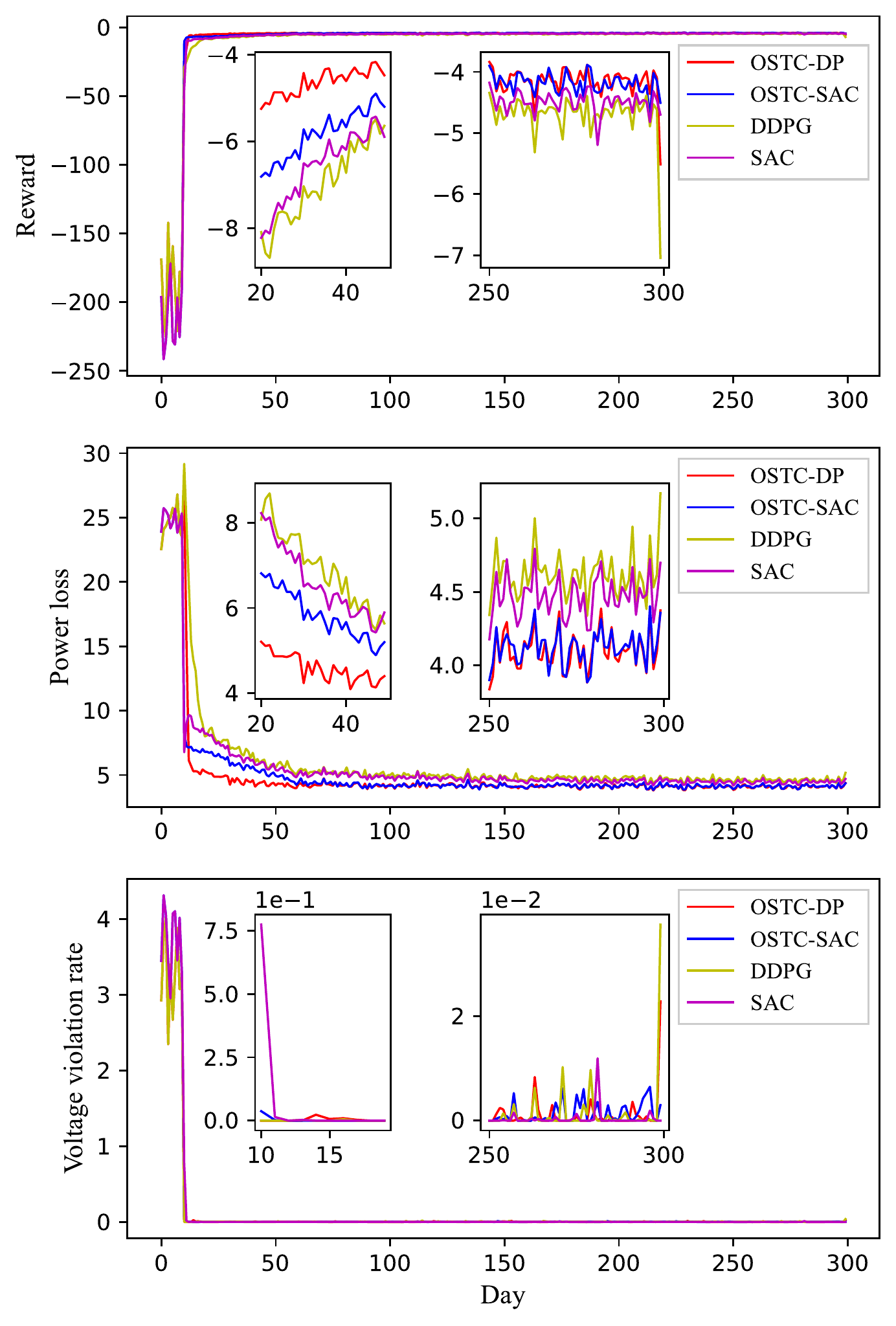}
\caption{Testing results of the training stage for the 69-bus test distribution network.}
\label{DRL_69}
\end{figure}


\begin{figure*}[htbp]\label{contribution}
\centering
\subfigure[The contribution of one-step]{\begin{minipage}[t]{0.38\linewidth}\label{contribution_os}
    \includegraphics[width=2.8in]{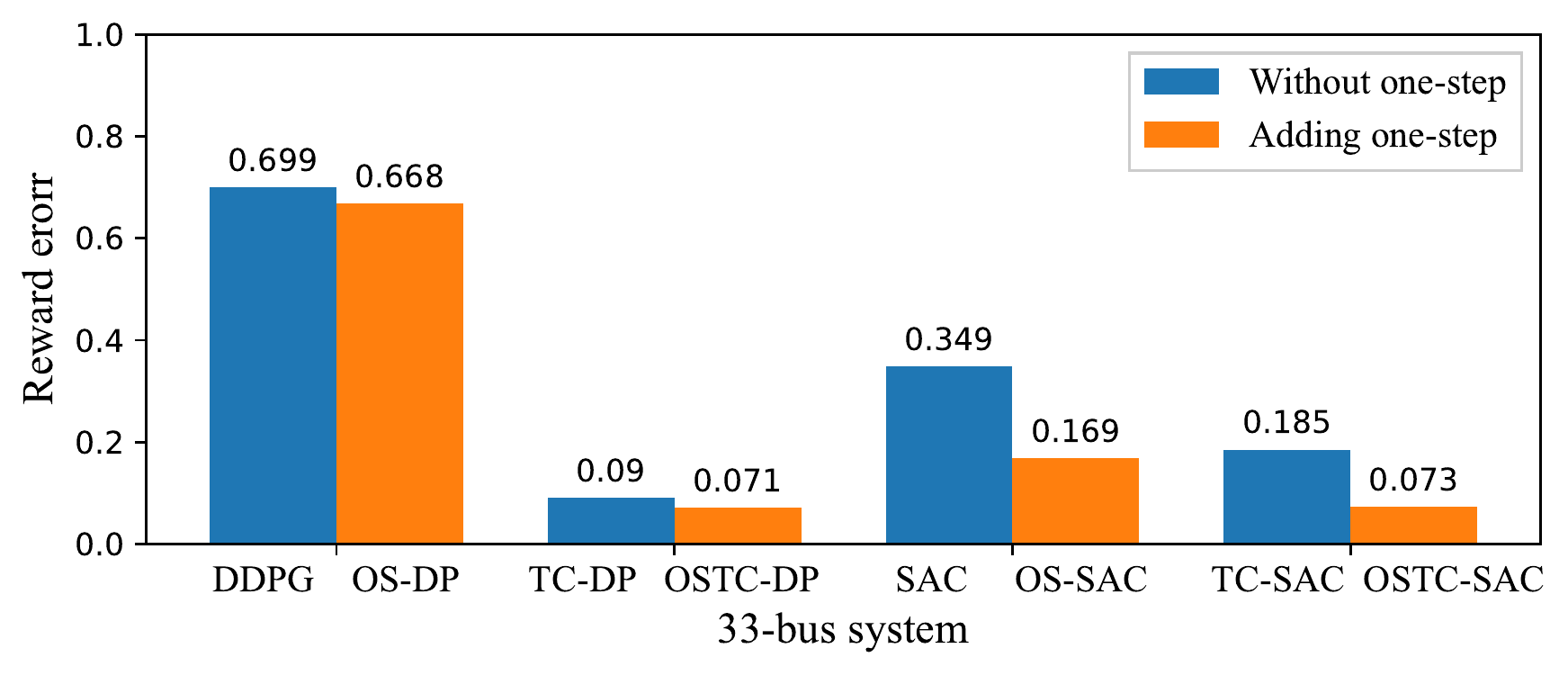}\\
    \includegraphics[width=2.8in]{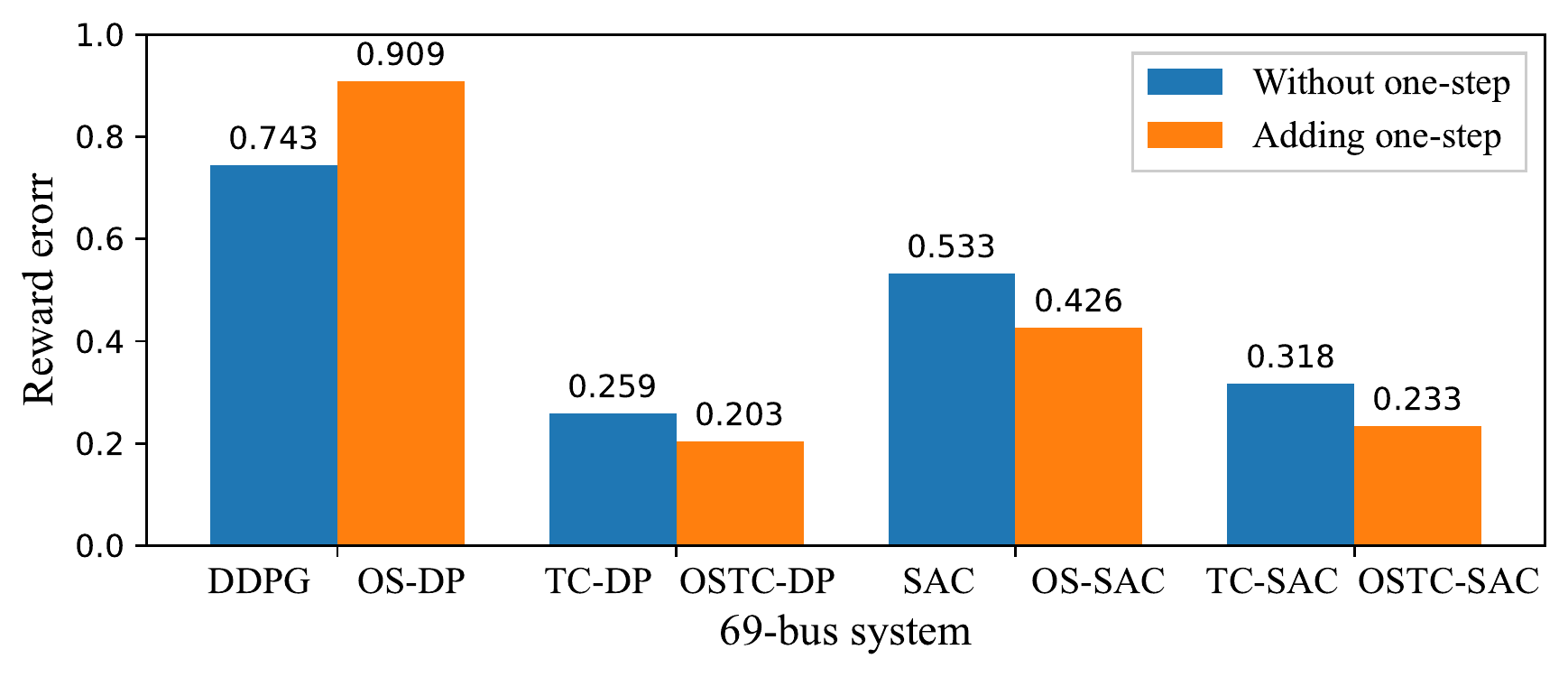}
    \end{minipage}
    }
\subfigure[The contribution of two-critic]{
\begin{minipage}[t]{0.38\linewidth}\label{contribution_tc}
\includegraphics[width=2.8in]{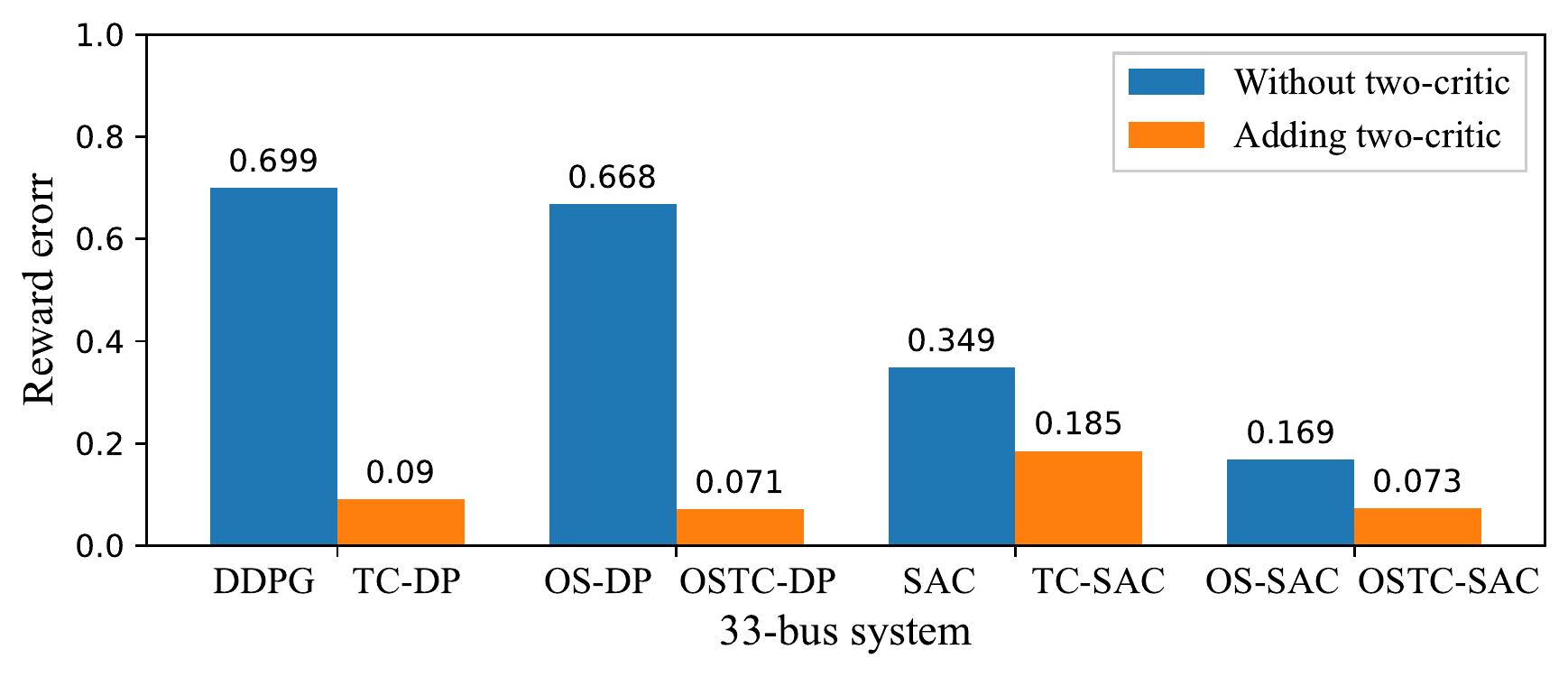}\\
\includegraphics[width=2.8in]{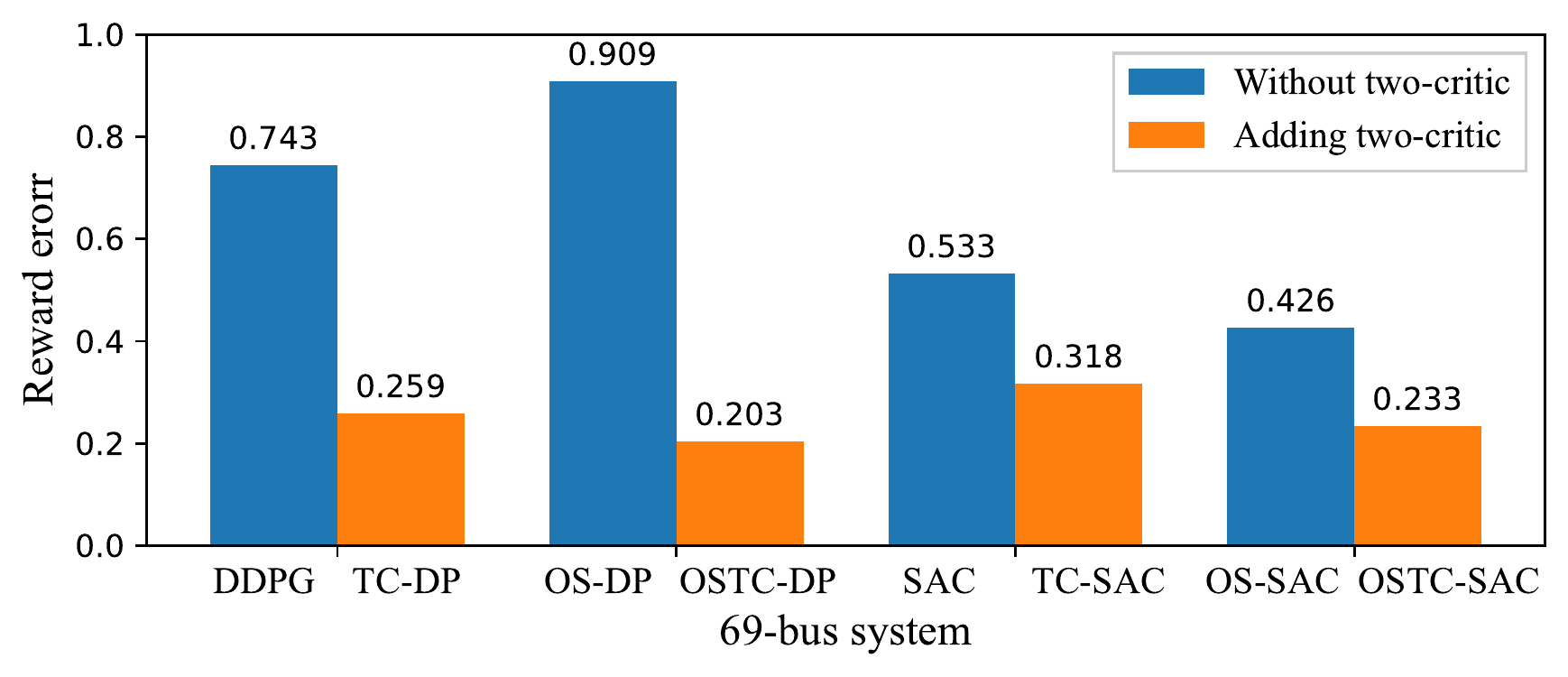}
\end{minipage}
    }
\subfigure[The contribution of one-step two-critic ]{\begin{minipage}[t]{0.20\linewidth}\label{contribution_ostc}
    \includegraphics[width=1.43in]{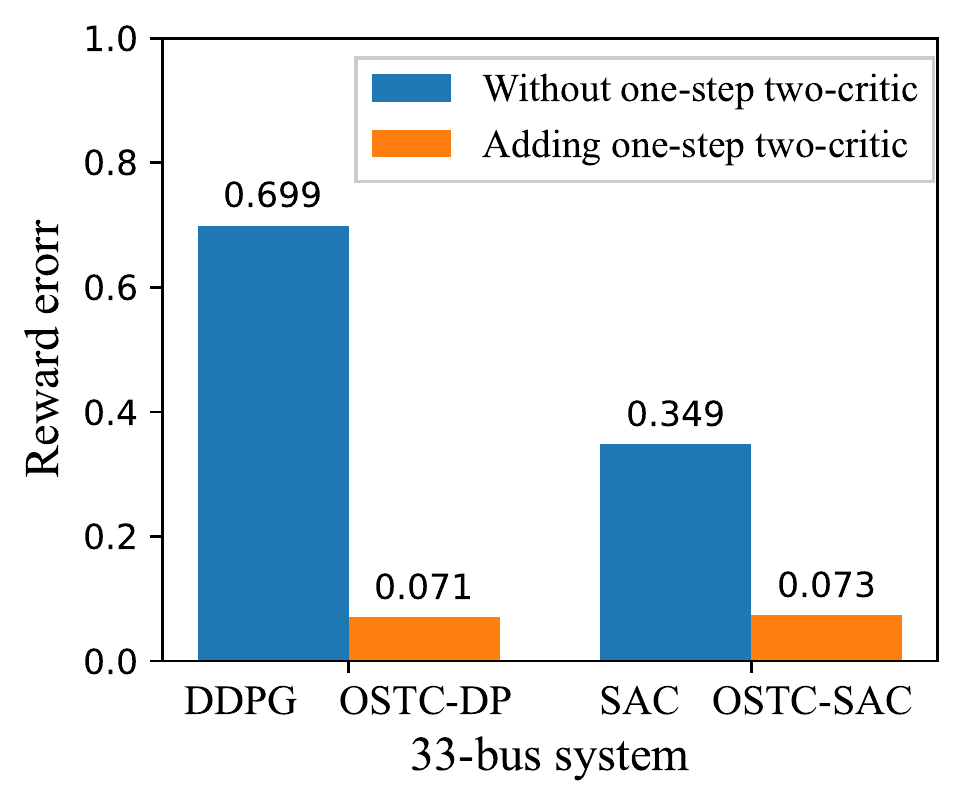}\\
    \includegraphics[width=1.43in]{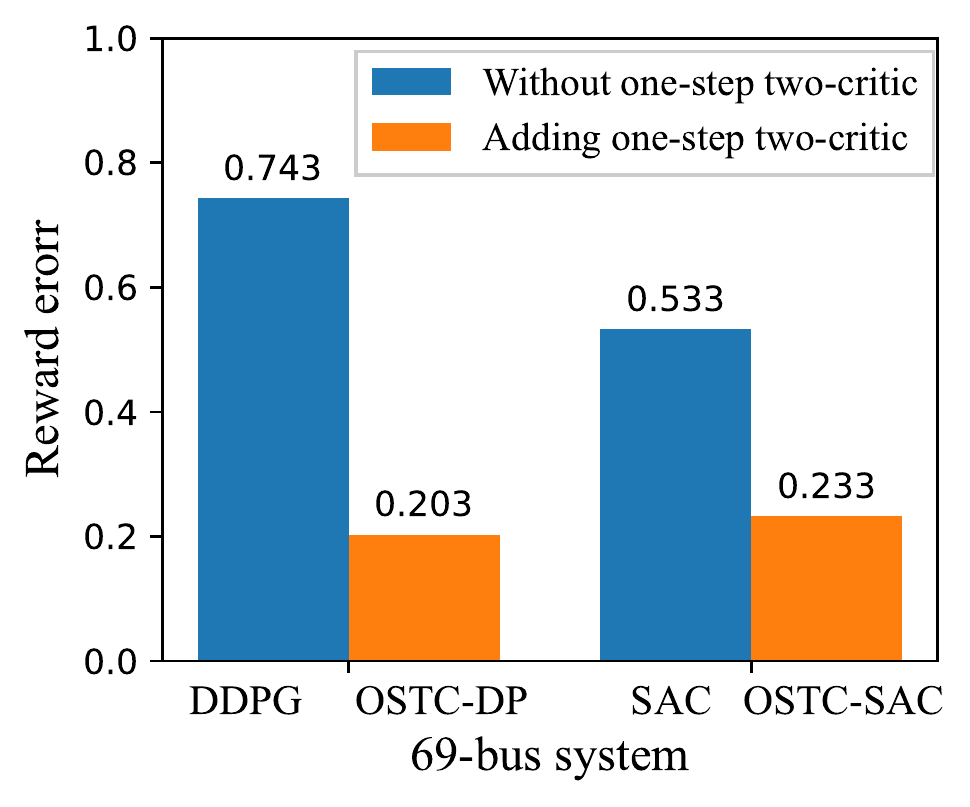}
    \end{minipage}
    }
\caption{The reward error in the final 50 episodes (reward error = reward of MBO - reward of the mentioned method) }\label{contribution}
\end{figure*}


The result of model-based optimization can be seen as optimal, which is a baseline for the performance of DRL algorithms.
We trained the DRL agent using 300 days of data. The parameter setting  for four DRL algorithms is provided in Table \ref{DRL_paramter}. The discount factor $\gamma$ for DDPG, TC-DP, SAC, TC-SAC is 0.9.
%
In the training process, we tested the DRL algorithms in the same environment at each step. 

The testing results in the training process are shown in Figs. \ref{DRL_33} and \ref{DRL_69}. The reward, power loss, and voltage violation rate in those figures are the daily accumulation values. 
To show the superiority of the proposed OSTC-DRL approach over the traditional DRL approaches clearly, we omitted the results of OS-DP, TC-DP, OS-SAC, and TC-SAC in Figs. \ref{DRL_33} and \ref{DRL_69}. We made the following two observations.

First, the proposed OSTC-DP and OSTC-SAC converged faster compared with DDPG and SAC, see the learning trajectory from days 20-50 in Figs. \ref{DRL_33} and \ref{DRL_69}.

Second, the proposed OSTC-DP and OSTC-SAC achieved higher rewards compared with DDPG and SAC, see the learning trajectory from  days 250-300 in Figs. \ref{DRL_33} and \ref{DRL_69}. OSTC-DP and OSTC-SAC also had smaller power losses.
	
The voltage violation rates of all DRL algorithms had  small fluctuations around zeros. Increasing the voltage violation penalty can decrease the voltage violation rate, but it just alleviates the fluctuations and cannot avoid voltage violation completely. DRL algorithms learn by trial and error, so in the training trajectory, DRL algorithms must trail both sides of the voltage boundary many times to find the optimal solution. 
The alternative way to address the voltage violation issues is by tightening the voltage limits. For example, for the normal voltage operation interval  $[0.95, 1.05]$, we set the objective voltage interval for DRL algorithms is $[0.955, 1.045]$. Even though there are slight voltage violations for the interval $[0.955, 1.045]$ in the learning process, there is no voltage violation for the interval $[0.95, 1.05]$. 

To quantify the advantages of the proposed OSTC-DRL approach, Table \ref{DRL_performance} gives the converged results of the 9 methods.
We use the accuracy equation $Acc = (R_{i}- R_{\text{MBO}})/R_{\text{MBO}} $ where $i$ represent OSTC-DP, DDPG, OSTC-SAC, or SAC, and MBO represents model based optimization method using an accurate power flow model.
From the perspective of reward, the accuracy of OSTC-DP, DDPG, OSTC-SAC, and SAC were $1.694\%$, $16.65\%$, $1.742\%$, and $8.304\%$ in the 33-bus network,  $5.133\%$, $18.79\%$, $5.891\%$, $13.47\%$ in the 69-bus network.
We can see that the accuracy of OSTC-DP was $9.831$ times as DDPG in the 33-bus network, and $3.660$ times as DDPG in the 69-bus network. The accuracy of OSTC-SAC was $4.766$ times as SAC in the 33-bus network, and $2.286$ times as SAC in the 69-bus network.
The results showed that the OSTC-DRL approach can improve the performance of DRL algorithms considerably for both deterministic policies and stochastic policies.

To show the contribution of the components of one-step and two-critic, Fig. \ref{contribution} gives the ablation study results. 
The contribution of the component of one-step is shown in Fig. \ref{contribution_os}.
The one-step improved the performance of those algorithms except that compares the results of DDPG and OS-DP for the 69-bus network.
The contribution of the component of the two-critic is shown in Fig. \ref{contribution_tc}.  Two-critic improved the performance of all of those algorithms.
The contribution of the combination of the one-step and the two-critic is shown in Fig. \ref{contribution_ostc}.
The algorithms with the OSTC-DRL approach achieved the best performance among the algorithms. 


Generally, SAC has better control performance than the DDPG because SAC has two extra components: clipped double-Q learning, and entropy regularization  \cite{haarnoja2018soft,haarnoja2018softsac}. 
However, our simulation results showed that the performances of TC-DP and OSTC-DP were better than the corresponding of TC-SAC and OSTC-SAC, respectively. The reasons may be as follows.
``clipped double-Q learning" mitigated the overestimation of Q value whereas bringing an underestimation bias \cite{ciosek2019better,pan2020softmax}. Entropy regularization accelerated the  learning process and prevented the policy from prematurely converging to a bad local optimum. However, it may bring additional regularization errors when the estimation accuracy of the Q function is high enough.
When the estimation accuracy of the  Q value was not  high, clipped double-Q learning and entropy regularization brought positive influence and led to the final results of SAC and OS-SAC over the corresponding DDPG and OS-DP. 
However, when the estimation accuracy of the  Q value was high, clipped double-Q learning and entropy regularization brought negative influence and led to the final results of TC-DP and OSTC-DP over the corresponding TC-SAC and OSTC-SAC, respectively.

\begin{table*}[tb]
\centering
\begin{threeparttable}
\renewcommand{\arraystretch}{1.3}
\caption{Quantified indices of the benchmarks in the final 50 episodes for  multi-agent DRL algorithms}
\label{MADRL_performance}
\centering
\begin{tabular}{cccccccc}
\hline 
 \multicolumn{2}{c}{\multirow{2}{*}{ Algorithm }} &  \multicolumn{2}{c}{Reward}  &  \multicolumn{2}{c}{$P_{\text {loss }} / \mathrm{MW}$}   & \multicolumn{2}{c}{VVR/p.u.}  \\
\cline { 3 - 8 } 
 & & 33-bus & 69-bus &  33-bus & 69-bus &  33 -bus & 69-bus \\
\hline
\multirow{3}{*}{\makecell[c]{Deterministic \\ policy}} 
		&OSTC-DP & $-4.270$ &  $-4.161$ & $4.268$ & $4.104$  & $3.988\text{e-}5$ & $1.137\text{e-}3$\\
        & MA-OSTC-DP-sub$^{1}$ & $-4.264$ & $-4.142$ & $4.262$ & $4.115$  & $2.787\text{e-}5$  & $5.364\text{e-}4$ \\
         & MA-OSTC-DP-local$^{2}$ & $-4.301$ & $-4.397$ & $4.288$ & $4.269$  & $2.624\text{e-}4$ & $2.568\text{e-}3$ \\
\hline
\multirow{3}{*}{\makecell[c]{Stochastic\\ policy}} &OSTC-SAC & $-4.272$ &$-4.191$ & $4.268$ & $4.121$& $9.205\text{e-}5$  & $1.402\text{e-}3$ \\
& MA-OSTC-SAC-sub & $ -4.343$ & $-4.234$ & $4.317$ & $4.128$ & $5.284\text{e-}4$  & $2.117\text{e-}3$ \\
& MA-OSTC-SAC-local & $-4.370$ & $-4.422$& $4.331$ & $4.282$ & $7.828\text{e-}4$  & $2.797\text{e-}3$ \\
\hline
\end{tabular}
\begin{tablenotes}
\item[1] ``MA" means multi-agent.  ''sub" represents the sub-area bus information. ``MA-OSTC-DP-sub'' means that the MA-OSTC-DP-sub algorithm works on that each actor executes based on the sub-area bus information.
\item[2]  ``local" represents the local one bus  information. ``MA-OSTC-DP-sub'' means that the MA-OSTC-DP-sub algorithm works on that each actor executes based on the local one bus information. 
\end{tablenotes}
\end{threeparttable}
\end{table*}



\subsection{Simulation for the Multi-Agent OSTC-DP Algorithm}

The proposed multi-agent algorithms were designed for measurement conditions of ADNs in that all the measurements are uploaded to the center at a slow rate and each agent only can obtain its sub-area measurements in real-time.
We tested the multi-agent algorithms on the two measurement conditions: 1) each agent can obtain the local one bus information, and  2) each agent can obtain the sub-area bus information. This setting was to show the multi-agent OSTC-DRL is flexible to different measurement situations.

For the first measurement conditions, the 33-bus test distribution network was divided into 4 subareas and the 69-bus test distribution network was divided into 5 subareas. Each subarea contained one bus in which the controllable devices have been installed.
For the second measurement condition, we divided the distribution networks into 4 sub-areas for both case 33 and 69 systems.
In the 33-bus test distribution network, the sub-areas were $[7,8,\dots, 18]$, $[19,20, 21,22]$, $[23, 24,25]$ and $[26,27,\dots, 33]$. In the 69-bus test distribution network, the sub-areas were $[2,3,\dots, 10]$, $[11,12,\dots, 26]$, $[36,37,\dots, 45]$ and $[53,54,\dots, 64]$. 
The partitioning is flexible. Some buses can belong to two partitionings concurrently, or not belong to any partitioning. 

Correspondingly, we extended OSTC-DP to multi-agent OSTC-DP-local and multi-agent OSTC-DP-sub for two measurement conditions. Also, we extended OSTC-SAC to multi-agent OSTC-SAC-local and multi-agent OSTC-SAC-sub. 
The parameters of DRL algorithms were the same as the corresponding subsection \ref{simulation_centralized} except for the number of actors. 
We tested the performance of DRL algorithms at each step in the training process.
Table \ref{MADRL_performance} shows the qualified indices of the 6 algorithms in the final 50 episodes.

After enough time to learn, for deterministic policies, multi-agent OSTC-DP-sub achieved similar performance as OSTC-DP, while the performance of  multi-agent OSTC-DP-local was slightly worse than the performance of OSTC-DP. 
For stochastic policies, the rank of the performance of three algorithms from high to low was OSTC-SAC, multi-agent OSTC-SAC-sub, and multi-agent OSTC-SAC-local. 
Noting that the performances of all multi-agent OSTC DRL algorithms are better than the performances of DDPG and SAC.
Those results showed that the multi-agent OSTC-DP and OSTC-SAC algorithms are robust for the information obtained by each actor in the execution stage, even when each actor just can obtain its local one bus information.
It is reasonable because the voltage information of one bus is influenced by other buses, so it can reflect the global information partially. 
Meanwhile, the partial information degrades the control performance of DRL slightly.

\section{Conclusion}
In this paper, we have proposed an OSTC-DRL approach for IB-VVC in ADNs.
Based on the OSTC-DRL approach, we designed two DRL algorithms that are OSTC-DP and OSTC-SAC.
We also extended the approach to the multi-agent OSTC-DRL approach for decentralized IB-VVC problems. 
We designed multi-agent OSTC-DP and multi-agent OSTC-SAC algorithms.
Simulation results showed the contributions of one-step and two-critic separately, and the OSTC-DRL approach has improved the VVC performance considerably compared with the state-of-the-art DRL algorithms for IB-VVC in ADNs.
After extending to the multi-agent DRL algorithms,
they achieved nearly equal or slight degradation performance as the OSTC-DRL algorithms.

The proposed OSTC-DRL approach focuses on single-period optimization problems and the action space is continuous. However, for ADNs embedded with capacity banks, on-load tap changers, and storage devices, the actions contain both continuous and discrete, and the optimization task should consider the long horizontal process. Therefore, we would extend our algorithm to mixed-integer multi-period optimization problems in future works.


%

\ifCLASSOPTIONcaptionsoff
  \newpage
\fi



\bibliographystyle{IEEEtran}
\bibliography{IEEEabrv,ref.bib}
\end{document}